\documentclass{article}

\usepackage[utf8]{inputenc} 
\usepackage[T1]{fontenc}    
\usepackage{hyperref}       
\usepackage{url}            
\usepackage{booktabs}       
\usepackage{authblk}
\usepackage{amsfonts}       
\usepackage{nicefrac}       
\usepackage{microtype}      
\usepackage{xcolor}         
\usepackage{mathtools}

\usepackage{tabularx}

\usepackage{times}
\usepackage{hyperref}
\usepackage{url}
\usepackage{multirow}
\usepackage{xr}

\usepackage{lipsum}
\usepackage{textcomp}

\usepackage{fontawesome}
\usepackage{bbm}
\usepackage{todonotes}
\usepackage{listings}
\usepackage{textgreek} 
\usepackage{amsmath} 
\usepackage{amssymb,amsthm,mathtools}

\usepackage[capitalise]{cleveref}
\usepackage{stfloats} 
\usepackage{float}
\usepackage{placeins} 

\usepackage[backend=biber,style=apa]{biblatex}
\usepackage{xcolor}

\usepackage{tablefootnote} 
\usepackage[roman]{parnotes} 
\usepackage{tabulary}
\usepackage{booktabs}
\usepackage{tabularx}

\usepackage{tikz}

\usetikzlibrary{shapes,decorations,arrows,calc,arrows.meta,fit,positioning}
\tikzset{
    -Latex,auto,node distance =1 cm and 1 cm,semithick,
    state/.style ={ellipse, draw, minimum width = 0.7 cm},
    point/.style = {circle, draw, inner sep=0.04cm,fill,node contents={}},
    bidirected/.style={Latex-Latex,dashed},
    el/.style = {inner sep=2pt, align=left, sloped}
}

\usepackage{changepage}

\usepackage{algorithm}  
\usepackage{algorithmicx}
\usepackage[noend]{algpseudocode}
\definecolor{mygray}{gray}{0.5}
\definecolor{cblue}{RGB}{8, 85, 153}
\definecolor{darkblue}{RGB}{31, 64, 96}

\definecolor{cgreen}{RGB}{8, 153, 83}
\definecolor{green}{RGB}{8, 200, 50}
\definecolor{cmaroon}{RGB}{128, 0, 0}
\algdef{SE}[DOWHILE]{Do}{doWhile}{\algorithmicdo}[1]{\algorithmicwhile\ #1}%

\renewcommand\algorithmicdo{}

\Crefname{section}{Sec}{Secs.}
\Crefname{figure}{Fig}{Figs.}
\Crefname{theorem}{Thm}{Thms.}

\usepackage{amsfonts}

\usepackage{tikzsymbols}

\usepackage[definitionLists,hashEnumerators,smartEllipses,hybrid]{markdown} 

\newcommand{\emptytree}{\tree_0}

\newcommand{\param}[2]{\textit{#1}={#2}}

\newcommand{\lklhd}{p}

\newcommand{\treespace}{\Omega}
\newcommand{\ncat}{m}
\newcommand{\nleaves}{b}

\newcommand{\bw}{\mathbf{w}}
\newcommand{\bx}{\mathbf{x}}
\newcommand{\by}{\mathbf{y}}

\newcommand{\bq}{\mathbf{q}}

\newcommand{\bX}{\mathbf{X}}

\newcommand{\bI}{\mathbf{I}}

\newcommand{\bTheta}{\boldsymbol{\Theta}}

\newcommand{\R}{\mathbb{R}}
\renewcommand{\P}{\mathbb{P}}
\newcommand{\E}{\mathbb{E}}

\newcommand{\sbart}{\text{simplified BART }}
\newcommand{\bcart}{\text{Bayesian CART}}

\newcommand{\Var}{\textnormal{Var}}

\DeclarePairedDelimiter{\braces}{\lbrace}{\rbrace}
\DeclarePairedDelimiter{\paren}{(}{)}

\DeclarePairedDelimiter{\abs}{|}{|}
\DeclarePairedDelimiter{\norm}{\|}{\|}

\newcommand{\tree}{\mathcal{T}}

\newtheorem{theorem}{Theorem}[]
\newtheorem{lemma}[theorem]{Lemma}

\title{A Mixing Time Lower Bound for a Simplified Version of BART}

\newcommand*\samethanks[1][\value{footnote}]{\footnotemark[#1]}

\author[1]{Omer Ronen\thanks{Equal contribution, alphabetical ordering}}
\author[1]{Theo Saarinen\samethanks}
\author[4]{Yan Shuo Tan\samethanks}
\author[6]{James Duncan}

\author[1,2,3,5]{Bin Yu}
\affil[1]{Department of Statistics, UC Berkeley}
\affil[2]{Department of Electrical Engineering and Computer Sciences, UC Berkeley}
\affil[3]{Center for Computational Biology, UC Berkeley}
\affil[4]{Department of Statistics and Data Science, National University of Singapore}
\affil[5]{Microsoft Research}
\affil[6]{Group in Biostatistics, UC Berkeley}

\begin{document}

\maketitle

\begin{abstract}
    Bayesian Additive Regression Trees (BART) is a popular Bayesian non-parametric regression algorithm. The posterior is a distribution over sums of decision trees, and predictions are made by averaging approximate samples from the posterior.
    The combination of strong predictive performance and the ability to provide uncertainty measures has led BART to be commonly used in the social sciences, biostatistics, and causal inference.
    BART uses Markov Chain Monte Carlo (MCMC) to obtain approximate posterior samples over a parameterized space of sums of trees, but it has often been observed that the chains are slow to mix.
    In this paper, we provide the first lower bound on the mixing time for a simplified version of BART in which 
    we reduce the sum to a single tree and use a subset of the possible moves for the MCMC proposal distribution.
    Our lower bound for the mixing time grows exponentially with the number of data points.
    Inspired by this new connection between the mixing time and the number of data points, we perform rigorous simulations on BART. We show qualitatively that BART's mixing time increases with the number of data points.
    The slow mixing time of the simplified BART suggests a large variation between different runs of the simplified BART algorithm and a similar large variation is known for BART in the literature. This large variation could result in a lack of stability in the models, predictions and posterior intervals obtained from the BART MCMC samples.
    Our lower bound and simulations suggest increasing the number of chains with the number of data points.
\end{abstract}

\section{Introduction}
\label{sec:intro}

Decision tree models such as CART \parencite{breiman1984classification} and their ensembles such as Random Forests \parencite{breiman2001random} and Gradient Boosted Trees \parencite{friedman2001greedy,chen2016xgboost} have proved to be enormously successful supervised learning algorithms, because they are able to combine non-parametric model fitting with implicit dimension reduction.
It is often difficult to quantify the uncertainty of their predictions and due to their greedy local splitting criteria, there is no guarantee for the optimality of the constructed decision trees.
An alternative approach is to construct the decision trees in a Bayesian manner \parencite{denison1998bayesian,chipman1998bayesian,wu2007bayesian}

To address these issues, \cite{chipman1998bayesian} proposed a Bayesian adaptation of CART, Bayesian CART, and later, a sum of Bayesian CART trees, which they called Bayesian Additive Regression Trees (BART) \parencite{chipman2010bart}.
One perspective views these algorithms as non-greedy stochastic versions of their deterministic equivalents, where the randomness inside the fitting process allows the algorithm to explore the space of possible decision trees in ways the CART algorithm cannot.
An alternative perspective views these algorithms as Bayesian non-parametric regression models, in which we put a prior on the space of decision trees, assume a likelihood for the observed data, and then obtain a posterior distribution over the possible decision trees based on the training data.
The posterior distribution can be used to provide posterior predictive credible intervals and other forms of uncertainty quantification.
Due to strong predictive performance \parencite{chipman2006bayesian,chipman2010bart} and the ability to quantify the uncertainty of predictions, BART has spawned a number of variants \parencite{hill2011bayesian,sparapani2016nonparametric,linero2018bayesian,hahn2020bayesian,pratola2020heteroscedastic,murray2021log} and has become increasingly popular in fields such as the social sciences \parencite{green2010modeling,yeager2019national}, biostatistics \parencite{wendling2018comparing,starling2020bart}, and causal inference \parencite{hill2011bayesian,green2012modeling,kern2016assessing,dorie2019automated,hahn2019atlantic,hill2020bayesian}.

Several recent works have theoretically analyzed BART variants from a frequentist perspective. 
Concentration results have been shown for the BART posterior under assumptions on the smoothness of the underlying regression function and the prior used for the BART algorithm.
\cite{rovckova2019theory} show that when the BART prior is modified to decrease the mass on deeper trees as the number of training points increases, the posterior distribution of this BART variant concentrates around the true regression function at nearly the optimal rate. 
When the true regression function exhibits smoothness, \cite{linero2018bayesian} introduce a smoothing variant of BART and show that the modified BART posterior concentrates at nearly the optimal rate for that class of functions.
In addition to concentration, there are several results for the feature selection consistency of the BART posterior.
When the regression function is smooth and sparse, \cite{rovckova2020posterior} show that BART with a sparsity inducing prior can perform effective feature selection.
\cite{liu2021variable} examine an approximate Bayesian computation algorithm to combine with BART for feature selection in high dimensional data.
However, all of these theoretical results rely on the ability to sample from the BART posterior.

\subsection{Prior Work on BART MCMC Mixing}
Since there is no closed form expression for the BART posterior, the standard approach is to sample from it approximately via a Markov chain Monte Carlo (MCMC) algorithm designed by \cite{chipman2010bart}.
Despite an abundance of empirical evidence, described following this, that the posterior samples from the BART algorithm do not mix well, researchers in many fields have regularly used the BART posterior for uncertainty quantification \parencite{green2012modeling,hill2013assessing,waldmann2016genome,bisbee2019barp,dorie2019automated,yeager2019national,zhang2020application,carlson2020embarcadero}.
Further, there has been very little theoretical analysis of how quickly the samples from the BART MCMC algorithm converge to the posterior distribution.
This is problematic for several reasons:
First, credible intervals obtained from the MCMC samples will not actually reflect the posterior, and are therefore of questionable meaning for inference.
Next, it means that different runs of the algorithm may produce somewhat different results, thereby lacking stability and reproducibility.
Finally, a necessary condition for the recent results on posterior concentration and model selection consistency, is the ability to sample from the posterior distribution. 
When the BART MCMC algorithm is slow to mix, it is difficult to satisfy this condition in practice.

Since the introduction of the BART algorithm, the problem of mixing has been observed,
\parencite{chipman2010bart,pratola2016efficient,he2021stochastic} leading to a number of works evaluating and attempting to address this issue.
The difficulties in mixing have been explored empirically in several works \parencite{chipman1998bayesian,wu2007bayesian,pratola2016efficient} and mentioned in several recent survey papers \parencite{linero2017review,hill2020bayesian}.
There have also been several algorithmic suggestions including modifying the MCMC proposal moves \parencite{pratola2016efficient,wu2007bayesian}, initializing the trees in the algorithm from greedily constructed trees as a "warm start" \parencite{he2021stochastic}, and running multiple chains to quantify the uncertainty in the predictions \parencite{dorie2019automated,Carnegie2019bartmixing}.
Despite the prevalence of works mentioning the mixing of the BART MCMC algorithm, little theoretical work has been done to understand why the mixing is slow. 

\subsection{BART and Bayesian CART}
In this paper we will discuss several variants of the BART algorithm.
The Bayesian CART algorithm was originally introduced as an algorithm for fitting a single decision tree in a Bayesian setting.
The BART algorithm fits a sum of these Bayesian CART trees and when this sum has one term, devolves to the Bayesian CART algorithm.
BART enjoys better prediction accuracy than Bayesian CART \parencite{chipman2006bayesian, chipman2010bart,hill2020bayesian}, and is hence used more often in practice.
Because Bayesian CART and each individual tree in the BART sum share the same set of MCMC proposal moves, Bayesian CART has been used to study the empirical effect of these proposal moves on the mixing of both algorithms \parencite{wu2007bayesian,chipman2010bart,pratola2016efficient,hill2020bayesian}.
In our theoretical work, we begin by using a simplified version of Bayesian CART towards the goal of explaining the properties of BART.
For theoretical tractability, we restrict the MCMC proposals available to the Bayesian CART algorithm to a subset of the standard moves and call the resulting algorithm \emph{simplified BART}. 
We present our theoretical mixing time lower bound for the simplified BART algorithm.
We compare the mixing time of BART to simplified BART in simulations in Section \ref{sec:expr1}, and compare Bayesian CART to simplified BART in Section \ref{sec:expr2} in order to compare the propensity of the two algorithms to select poor initial splitting features.
\begin{table}[H]
    \centering
    \caption{A summary of the different algorithms studied in this paper.}
    \vspace{\baselineskip}
    \resizebox{\columnwidth}{!}{
\begin{tabular}{lrrrr}
\toprule
       Algorithm \textbackslash\;Property  & Number of Trees &  MCMC moves \\
\midrule
       BART &   200   & \textbf{Grow}, \textbf{Prune}, \textbf{Change}, \textbf{Swap}\\
 Bayesian CART & 1 & \textbf{Grow}, \textbf{Prune}, \textbf{Change}, \textbf{Swap}\\
 Simplified BART & 1 &  \textbf{Grow}, \textbf{Prune}\\
\bottomrule
\end{tabular}}
    \label{tab:bart_variants}
\end{table}

\subsection{Our contributions}

We present a theoretical result for the failure of the simplified BART MCMC algorithm to mix and show through extensive simulations that the BART MCMC algorithm has a similar mixing issue. As far as we know, our work is the first attempt at theoretically analyzing the mixing behavior of a simplified version of the Bayesian CART or BART algorithms, with insights that also hold qualitatively for BART.
Our specific contributions are as follows:
\begin{enumerate}
\item \textbf{Mixing time lower bounds:}
We obtain a mixing time lower bound for a simplified version of the BART algorithm. 
Assuming that all features are discrete, we are able to show that the total variation mixing time of the simplified BART MCMC chain on the space of tree structures is bounded from below by $\exp(\Omega(n))$, where $n$ is the number of training data points.
This theoretical result holds for any data generating process with features from any random distribution on a discrete feature space of dimension at least $2$ and a random outcome vector $y$.
In addition, our proof identifies one reason for the slow mixing:
It is difficult for the chain to switch the split at the root of the tree.

\item \textbf{Simulations:}
Taking several large real datasets from the Penn Machine Learning Benchmark repository (PMLB) \parencite{olson2017pmlb}, we run 8 independent instances of the MCMC chain for BART and for simplified BART and study the mixing of the chains according to several metrics.
We use the root mean squared error (RMSE) from a held out test set as a summary statistic of each sampled regression function, and compute the distribution of this quantity over each chain after discarding a conservative number of burn-in samples.
The failure of the chain to mix can be observed visually by comparing the density plots for the RMSE of the different chains (see \cref{fig:bart_rmse}).
We also compute the Gelman-Rubin (GR) diagnostic values across the different chains, and observe that in many cases these exceed the threshold of 1.1 that was recommended by \cite{gelman1992inference} as a quantitative indicator of the failure of mixing. 
Through our simulations, we find that \emph{the BART algorithm shows worse mixing as the number of training data points increases}, as indicated by mixing diagnostics as well as qualitative measures of the regression function.
We also find that Bayesian CART is susceptible to the bottleneck that we exploit for simplified BART in the proof of our theoretical result: Difficulty in reversing the split at the root of the tree. We did not find strong evidence that this bottleneck affects the BART algorithm to the same degree.
\end{enumerate}
To the extent of our knowledge, our theoretical result is the first that studies the mixing time of either the BART or Bayesian CART algorithms.
Our simulations suggest that the problems identified theoretically for simplified BART still affect the original BART and Bayesian CART algorithms.
In particular, the issue of mixing time increasing with the number of data points suggests that BART practitioners may benefit by running more chains for data with many observations, an observation not yet explored in the literature.

\section{Preliminaries}

\label{sec:methods}

In this section, we describe the elements of the Bayesian CART model underlying the simplified BART algorithm as well as the MCMC sampling algorithm, as described in the original paper \parencite{chipman1998bayesian}.
We define the Bayesian CART model formally because it only differs from the simplified BART algorithm we analyze in the set of MCMC moves.

\subsection{The Bayesian model specification}
\label{subsec:bayesian_model}

\paragraph{Parameter space:}
Recall that the Bayesian CART model is a non-parametric regression model, which means that the regression function $f$ is a random function.
Unlike Gaussian process regression, we constrain it to take values on the space of decision tree functions.
We say that $f\colon \mathcal{X}^d \to \R$ is a decision tree function if it is a piecewise constant function on a partition of $\mathcal{X}^d$, where the partition is generated by recursively splitting $\mathcal{X}^d$ along the coordinate directions.
Naturally, $f$ can be parameterized by a tuple comprising the binary tree structure $\mathcal{T}$, and $\bTheta \in \R^b$, where $b = \abs{\mathcal T}$ is the number of leaves in $\mathcal{T}$, and $\bTheta_i$ is the value of $f$ on leaf $i$ (after choosing a canonical ordering of the leaves).
In turn, the tree structure $\mathcal{T}$ can be thought of as a labeled graph, and thus further parameterized by the underlying ordered rooted binary tree, as well as labels on each internal node of the form $(v_j, \tau_j)$, which respectively denote the feature and threshold for the split associated with node $j$.
We assume that our covariate space is discrete, i.e. $\mathcal{X} = \{1,\ldots,\ncat\}$ for some integer $\ncat$.\footnote{This is almost WLOG as in practice, splits for continuous features are chosen from a grid of possible values corresponding to the quantiles of the features in the training data.}
This implies that a tree structure $\mathcal T$ can have at most $\ncat^d$ leaves, which, together with a finite choice of labels at each node, implies that $\treespace$, the collection of all tree structures, is discrete and finite.

\paragraph{Prior for $\mathcal T$:}
The prior on the tree structure is defined in terms of a stochastic process:
Starting with a trivial tree with a single node, each newly generated node is split with probability $\alpha(1+\delta)^{-\beta}$, where $\delta$ is the depth of the node, while $\alpha$ and $\beta$ are universal hyperparameters.
If a node is split, the feature it splits on is drawn uniformly from all available features, and then the threshold is drawn uniformly from all available values of the feature, if it is discrete, and from all available values that have been observed in the training data, if it is continuous.

\paragraph{Prior for $\bTheta$:}
We put independent Gaussian priors for the leaf parameters: $\bTheta|\mathcal T \sim \mathcal N(\bar\mu \textbf{1}, \sigma^2 \textbf{I}_b)$.
In the original Bayesian CART algorithm, $\sigma^2$ is treated as a Bayesian parameter drawn from an inverse Gamma distribution.
For the sake of theoretical tractability, we shall treat it instead as a fixed hyperparameter.
We believe that this is relatively innocuous as, in practice, $\sigma^2$ is known to quickly concentrate around a fixed value.

\paragraph{Data likelihood:}

We put an independent Gaussian likelihood function on the errors in the responses.
In other words, given observed data $\bX = \left(\bx_i \right)_{i=1}^n$, $\by = (y_i)_{i=1}^n$, we assume
$$
\by|\bX,\Theta,\mathcal T \sim \mathcal{N}\paren*{(f(\bx_i))_{i=1}^n, a\sigma^2\textbf{I}_n},
$$
where $a > 0$ is a fixed number.

The relationships between the variables described in this section can be summarized in the following Bayesian network diagram.
\begin{figure}[H]
    \centering
    \includegraphics[scale=0.4]{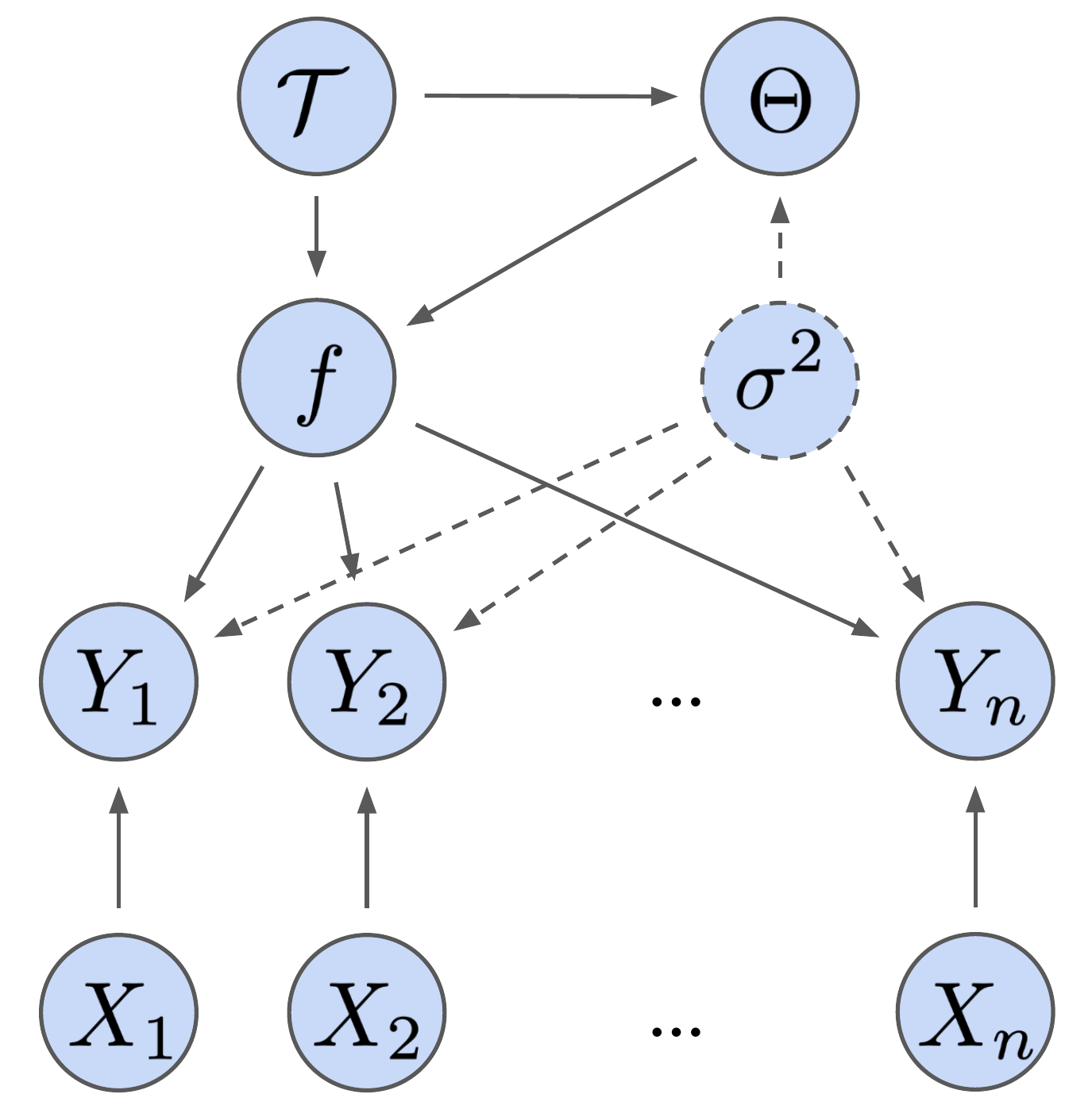}
    \caption{A Bayesian network showing the dependency relationships between the random variables in the simplified BART model.
    $\sigma^2$ is treated as a parameter in Bayesian CART, but will be treated as a fixed hyperparameter in our theoretical analysis.}
    \label{fig:bayesian_diagram}
\end{figure}

\subsection{Sampling from the Bayesian CART posterior}
\label{subsec:sampling}

To sample from the posterior $p(\mathcal T, \bTheta|\bX,\by)$, we first decompose it as $p(\bTheta|\mathcal T,\bX,\by)p(\mathcal T|\bX,\by)$.
The Gaussian specification of $\Theta$ and the data likelihood allow us to compute the first term in closed form, and so, we only need to use MCMC to sample from $p(\mathcal T|\bX,\by)$, the marginal posterior on the space of tree structures.
We do this using the Metropolis-Hastings algorithm.
When the chain is at a given tree $\mathcal T$, the proposed next tree $\mathcal T'$ is obtained by randomly choosing from one the following four moves: 
1. \textbf{Grow} the tree by splitting a leaf node chosen uniformly at random.
The split feature and threshold are also chosen uniformly at random, as in the prior.
2. \textbf{Prune} the tree by collapsing a pair of adjacent leaf nodes chosen uniformly at random.
3. \textbf{Change} the split feature and threshold (draw them again from the uniform distribution) of an internal node selected uniformly at random.
4. \textbf{Swap} the split features and thresholds of a parent-child node pair chosen uniformly at random.
An accept-reject filter is then applied to ensure that the chain has the marginal posterior as the stationary distribution.
The original simplified BART paper proposed selecting each move with equal probability.
Later when proposing BART, the authors updated their suggestion to use the probabilites $(0.25, 0.25, 0.4, 0.1)$.

\subsection{Simplified BART}
\label{subsec:simplified_bart_def}
We define the simplified BART algorithm as the Bayesian CART algorithm defined in Section \ref{subsec:sampling} with the MCMC moves restricted to \textbf{Grow} and \textbf{Prune} each of which we select with probability $.5$.
This is the algorithm that we theoretically analyze in Section \ref{sec:lb}.

\subsection{Contrasting Bayesian CART with BART}
\label{sec:bart_vs_bcart}
BART is Bayesian model for a tree sum that is built on top of Bayesian CART.
The same priors on tree structure and leaf parameters are used, while the likelihood function of the noise in the response remains Gaussian. However, the regression function is now the sum of all the tree functions in the sum, and the parameter space thus comprises the product $\paren*{\mathcal{T}_i, \bTheta_i}_{i=1}^M$, where $M$ is the number of trees.
To sample from the resulting posterior, we use a combination of Gibbs sampling and the Metropolis-Hastings algorithm. 
More precisely, we cycle through the trees in the model and update them as follows:
Given a tree $\tree_i$ in the model, conditioned on the values of parameters from all the other trees, we update $\tree_i$ using the same procedure as in Bayesian CART, except that we replace the responses with residuals in the likelihood function. 
We then make a single draw from the conditional posterior for $\bTheta_i$, and then repeat the process for the next tree in the sum.
A single update of the MCMC chain comprises an update for all $M$ trees in the sum.

\section{Mixing Time Lower Bound}
\label{sec:lb}

\paragraph{Notation on probabilities:}
We will use $p(-)$ to denote the marginal and conditional probabilities associated with the Bayesian model described in \cref{subsec:bayesian_model}.
We use $Q(-)$ to denote probabilities associated with the Markov chain for the simplified BART algorithm on $\treespace$, the space of tree structures described in \cref{subsec:sampling}.
As a shorthand, we will also denote the stationary distribution, the posterior marginal on $\Omega$, as $\pi$.
Finally, we assume that our observed data $(\bX, \by)$ comprise $n$ i.i.d. data points from a generative regression model with regression function $f_0(\bx) = \E\braces{y~|~\bx}$.
We will denote probabilities, expectations, and variances with respect to the generative process using $\P$, $\E$, and $\Var$.
Note that the generative model is not necessarily the same as the data likelihood in the fitted Bayesian model.

The mixing time (see also \cite{LevinPeresWilmer2006}) of the Markov chain of the simplified BART algorithm, $Q$, is defined to be
\begin{align}
\label{def:sbart_tmix_def}
    t_{mix} \coloneqq \min\{t~\colon~\max_{\tree \in \treespace}\Vert Q^t(-|\tree) - \pi \Vert_{\text{TV}}\leq 0.25\}.
\end{align}
The quantity being maximized over in the right hand side of the equation is the total variation distance between the time $t$ distribution of the chain initialized at a tree structure $\tree$ and the stationary distribution.
A larger mixing time means that the chain takes a long time to reach the stationary distribution from a worst case initialization.\footnote{The choice of 0.25 as the threshold is by convention and does not affect the mixing time up to multiplicative constants.}

The main theoretical result of our paper is the following theorem.
\begin{theorem}[Mixing time lower bound]
    \label{thm:main_thm}
    Suppose $d \geq 2$ or $\ncat \geq 2$.
    Also assume that $y$ has a bounded distribution, i.e. $\abs{y} \leq K$.
    Then with probability at least $1-1/n$, the mixing time of the simplified BART Markov chain, $Q$, as described in \cref{{subsec:simplified_bart_def}}, satisfies
    \begin{equation} \label{eq:mixing_time_bound_final}
        t_{mix} \geq \exp\left(\frac{n}{2a\sigma^2}\Var\lbrace f_0(\bx)\rbrace - O(\sqrt{n\log n})\right).
    \end{equation}
\end{theorem}

The theorem says that under trivial assumptions, the mixing time of the simplified BART MCMC algorithm grows exponentially in the number of data points.
This is surprising and unfortunate because we generally expect the performance of prediction algorithms to improve with more data points.

Our proof of the lower bound proceeds by constructing two tree structures $\tree$ and $\tree'$ that each give rise to a partition on which $f_0$ is piecewise constant, but whose splits at the root node differ from each other.
This means that both $\tree$ and $\tree'$ possess large posterior mass, but are separated from each other in the state space by a bottleneck for the chain -- the trivial tree with only a root node.
This bottleneck becomes increasingly difficult to traverse as the number of data points increases.

While the result concerns mixing for the tree structure $\tree$ and not the regression function $f$, in most cases we may select $\tree$ and $\tree'$ so that the partition associated with $\tree$ is a refinement of that associated with $\tree'$.
In this case, the resulting conditional posterior distributions $p(f~|~\tree,\bX,\by)$ and $p(f~|~\tree',\bX,\by)$ are different, which
strongly suggests that the stochastic process induced by $Q$ on the space of regression functions\footnote{The transition kernel on the full parameter space is given by $\tilde Q(\tree',\bTheta'|\tree,\bTheta) = Q(\tree'|\tree)p(\bTheta'|\tree',\bX,\by)$.
Since the tuple $(\tree, \bTheta)$ determines $f$, this chain induces a stochastic process on the space of regression functions.} 
also mixes slowly with respect to Wasserstein distances.
Such a result, while even more revealing, would be relatively technically difficult, and we leave it to future work.

The notion of a bottleneck is formalized by the definition of \emph{conductance}.
The conductance of a subset $S \subset \Omega$ is defined as the ratio
\begin{equation*}
    \Phi(S) \coloneqq \frac{Q(S,S^c)}{\min\braces{\pi(S),\pi(S^c)}},
\end{equation*}
where
\begin{align*}
    Q(S, S^c) = \sum_{\tree \in S, \tree' \in S^c}\pi(\tree)Q(\tree'| \tree)
\end{align*}
is the probability of starting in $S$ and ending in $S^c$ over one step of $Q$ when applied to the stationary distribution.
A small conductance implies that it is difficult for $Q$ to transit from $S$ to $S^c$.
This intuition can be used to derive the following well-known inverse relationship between mixing time and conductance.

\begin{lemma}[Conductance and mixing time, Theorem 7.3 in \cite{LevinPeresWilmer2006}]
\label{lma:conductance_mixing_time}
    \begin{equation*}
        t_{mix} \geq \frac{1}{4\Phi^*}
    \end{equation*}
    where
    $t_{mix}$ is the mixing time of the simplified BART algorithm defined in \ref{def:sbart_tmix_def}
    and
    \begin{equation*}
        \Phi^* \coloneqq \min_{S \subset \Omega} \Phi(S).
    \end{equation*}
\end{lemma}
By using the bottleneck at the trivial tree alluded to earlier, we can upper bound the conductance in terms of the minimum of two likelihood ratios, each of which compares the trivial tree to a nontrivial one making a different root split.

\begin{lemma}[Conductance and likelihood ratio]
\label{lma:conductance_lratio}
Let $\tree$ and $\tree'$ be two tree structures whose splits at the root node differ from each other.

Then
\begin{equation} \label{eq:conductance_bound_final}
    \Phi^* \leq C \min\braces*{\frac{\lklhd(\by|\tree_0,\bX)}{\lklhd(\by|\tree,\bX)}, \frac{\lklhd(\by|\tree_0,\bX)}{\lklhd(\by|\tree',\bX)}},
\end{equation}
where $C = C(\tree,\tree',\alpha,\beta,m,d)$ is a constant not depending on $n$, and $\tree_0$ is the trivial tree comprising only a root node.
\end{lemma}
To further bound each likelihood ratio in \eqref{eq:conductance_bound_final}, we formulate a more general result for the log likelihood ratio between the trivial tree and \emph{any} candidate tree $\tree$.
The result shows that the normalized log likelihood ratio concentrates around the population \emph{impurity decrease} between $\tree_0$ and $\tree$.

\begin{lemma}[Likelihood ratio and impurity decrease]
\label{lma:lratio_impurity}
    For any tree $\tree$, with probability at least $1-1/n$, we have
    \begin{align*}
        & \abs*{\log\frac{\lklhd(\by|\tree_0,\bX)}{\lklhd(\by|\tree,\bX)} + \frac{n\Delta}{2a\sigma^2}} \\
        \leq~ &  CK^2a^{-1}\sigma^{-2}\sqrt{\log n}\paren*{\sqrt{n} + b} + b\log(n/a).
    \end{align*}
    where
    \begin{equation*}
        \Delta \coloneqq \Var\braces*{f_0(\bx)} - \E\braces*{\Var\braces*{f_0(\bx)~|~\tree(\bx)}}
    \end{equation*}
    is the decrease in impurity of $f_0(\bx)$ when conditioning on the partition given by $\tree$, and $C$ is an absolute constant.
\end{lemma}

The empirical impurity decrease is used by CART to choose which splits to make and which to prune, while the likelihood ratio is used by Bayesian CART to accept or reject proposed grow, prune, change, or swap moves.
This lemma thereby further illustrates the similarities and differences between the two algorithms.
More importantly for this paper, it completes the toolbox we need to prove our main theorem.

\begin{proof}[Proof of Theorem 1]

By assumption, there are at least two possible splits at the root node, $(v,\tau)$ and $(v',\tau')$.
Starting from either split, it is possible to grow a tree on whose leaves the true regression function is piecewise constant.
Call these two trees $\tree$ and $\tree'$ respectively, and notice that they automatically satisfy the hypotheses of Lemma \ref{lma:conductance_lratio} so that \eqref{eq:conductance_bound_final} holds.
Furthermore, we have
\begin{equation*}
    \Var\braces*{\E\braces{f_0(\bx)|\tree(\bx)}} = \Var\braces*{\E\braces{f_0(\bx)|\tree'(\bx)}} = 0.
\end{equation*}
By Lemma \ref{lma:lratio_impurity}, we therefore have
\begin{equation*}
    \log\frac{\lklhd(\by|\tree_0,\bX)}{\lklhd(\by|\tilde\tree,\bX)} = - \frac{n}{2\sigma^2}\Var\braces*{f(\bx)}  + O(\sqrt{n\log n} )
\end{equation*}
for $\tilde\tree = \tree, \tree'$.
Exponentiating and applying Lemma \ref{lma:conductance_lratio} followed by  Lemma \ref{lma:conductance_mixing_time} then gives us \eqref{eq:mixing_time_bound_final}.
\end{proof}

\externaldocument{07_additional_sims}

\section{Simulations on Mixing for BART and simplified BART}
\label{sec:sims}

In this section, we perform two simulation experiments to understand how the conclusions from our theoretical analysis of \sbart in the previous section provide new insights for understanding BART and Bayesian CART. 
Specifically our simulations suggest that:
\begin{itemize}
    \item The BART MCMC algorithm often mixes poorly, and the mixing quality decreases as the number of training data points $n$ increases.
    \item The root split made by the \bcart~algorithm is often chosen suboptimally and yet is rarely reversed, thereby creating a bottleneck for the chain.
\end{itemize}

\label{sec:sim_code}
\paragraph{Code Availability:}
All the code necessary to reproduce the experiments in this section is publicly available at \href{https://github.com/theo-s/bart-sims}{\faGithub\,github.com/theo-s/bart-sims}
The computing infrastructure used was a Linux cluster managed by Department of Statistics at UC Berkeley. Most runs of the simulation used a single 24-core node with 128 GB of RAM, while the larger datasets required a large-memory node with 792 GB RAM and 96 cores.

\label{sec:sim_data}
\paragraph{Data:}
We use real-world datasets in order to emulate how BART 
is used in practice. 
Specifically, we utilize the four largest datasets from the Penn Machine Learning Benchmarks (PMLB) \parencite{olson2017pmlb} in order to study how the mixing time depends on the number of training data points.
Table \ref{tab:datasets} details the dimensions of these datasets.

\begin{table}[H]
    \centering
    \caption{PMLB datasets utilized in simulations}
    \vspace{\baselineskip}
    \resizebox{\columnwidth}{!}{
    \begin{tabular}{lrr}
\toprule
                                     Name &  Samples &  Features \\
\midrule
       Breast tumor \parencite{romano2020pmlb} &   116640 &         9 \\
 California housing \parencite{pace1997sparse} &    20640 &         8 \\
        Echo months \parencite{romano2020pmlb} &    17496 &         9 \\
    Satellite image \parencite{romano2020pmlb} &     6435 &        36 \\
\bottomrule
\end{tabular}}
    \label{tab:datasets}
\end{table}

\paragraph{Algorithm settings and hyperparameters:}
We use the \textbf{dbarts} R package \parencite{dbarts} with the following non-default hyperparameters. First, we increase the number of burn-in samples from 100 to 5000 (\param{nskip}{5000}), in order to highlight that mixing does not occur within a reasonable number of iterations. Second, we run 8 chains (\param{nchain}{8}) to facilitate the analysis of the mixing time. Last, we define the proposal distribution of \sbart to select grow and prune with equal probability (\param{probs}{(1-1e-5,1e-5,0,.5)}\footnote{1e-5 is used avoid a numeric error, this term does not affect the proposal probabilities.}). All other hyperparameters are kept at their default values. In particular, for simplified BART we use \param{ntree}{1} and for BART we use \param{ntree}{200}.

\subsection{Experiment 1: Mixing time increases with n}
\label{sec:expr1}

\paragraph{Choice of mixing metric:}
Each sample in a chain is a function, and we calculate its RMSE on a held-out evaluation set, comprising 10\% of the dataset, as a one-dimensional summary statistic. 
To quantify the degree of mixing, we compute the Gelman-Rubin convergence diagnostic (GR) \parencite{gelman1992inference}, which compares the within-chain variation to the across-chain variation.\footnote{The Gelman-Rubin diagnostic is a statistic that uses the observation that multiple MCMC chains should realize similar values under convergence to give a measure of convergence for a collection of MCMC chains. Suppose we have $L$ samples from $J$ different MCMC chains, let $x_{ij}$ denote the $i$th sample from the $j$th chain, $\bar x_j$ denote the mean sample from the $j$th chain, and $\bar x$ denote the overall mean sample. The Gelman-Rubin diagnostic is defined as the ratio $R = \frac{\frac{L-1}{L} W+\frac{1}{L} B}{W}$ where $W$ denotes the within-chain variance: $W := \frac{1}{J} \sum_{j=1}^J \frac{1}{L-1} \sum_{i=1}^L\left( x_{ij} - \bar x_j \right)^2$ and $B$ denotes the between-chain variance: $B := \frac{L}{J-1} \sum_{j=1}^J\left(\bar x_j - \bar{x} \right)^2$.}  
It is widely accepted that a GR value which exceeds 1.1 indicates failure to mix \parencite{gelman1992inference}.
We implicitly assume that the degree of mixing at a fixed number of iterations is indicative of the mixing time.

\paragraph{Varying $n$:}
To vary $n$, we draw random sub-samples without replacement from each dataset of sizes $n=200$ and $n=2000$ in addition to using the full dataset.

\begin{figure}[t]
    \includegraphics[width=0.9\textwidth]{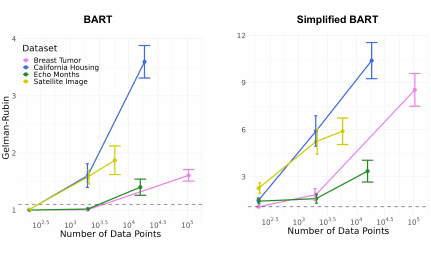}
    \centering
            \caption{Mixing quality for BART and \sbart decreases with the number of data points. 
        Mixing quality is measured in terms of the Gelman-Rubin diagnostic applied to the test set RMSE using 8 chains.
        Error bars are calculated over 20 different Monte carlo runs using the same training data. 
        The left column shows the results for BART, while the right for \sbart.
        The horizontal dashed line represent the recommended mixing threshold of $1.1$.
        }
    \label{fig:gr_n_p_bart}
\end{figure}

\begin{figure}[H]
    \centering
    \includegraphics[scale=0.8]{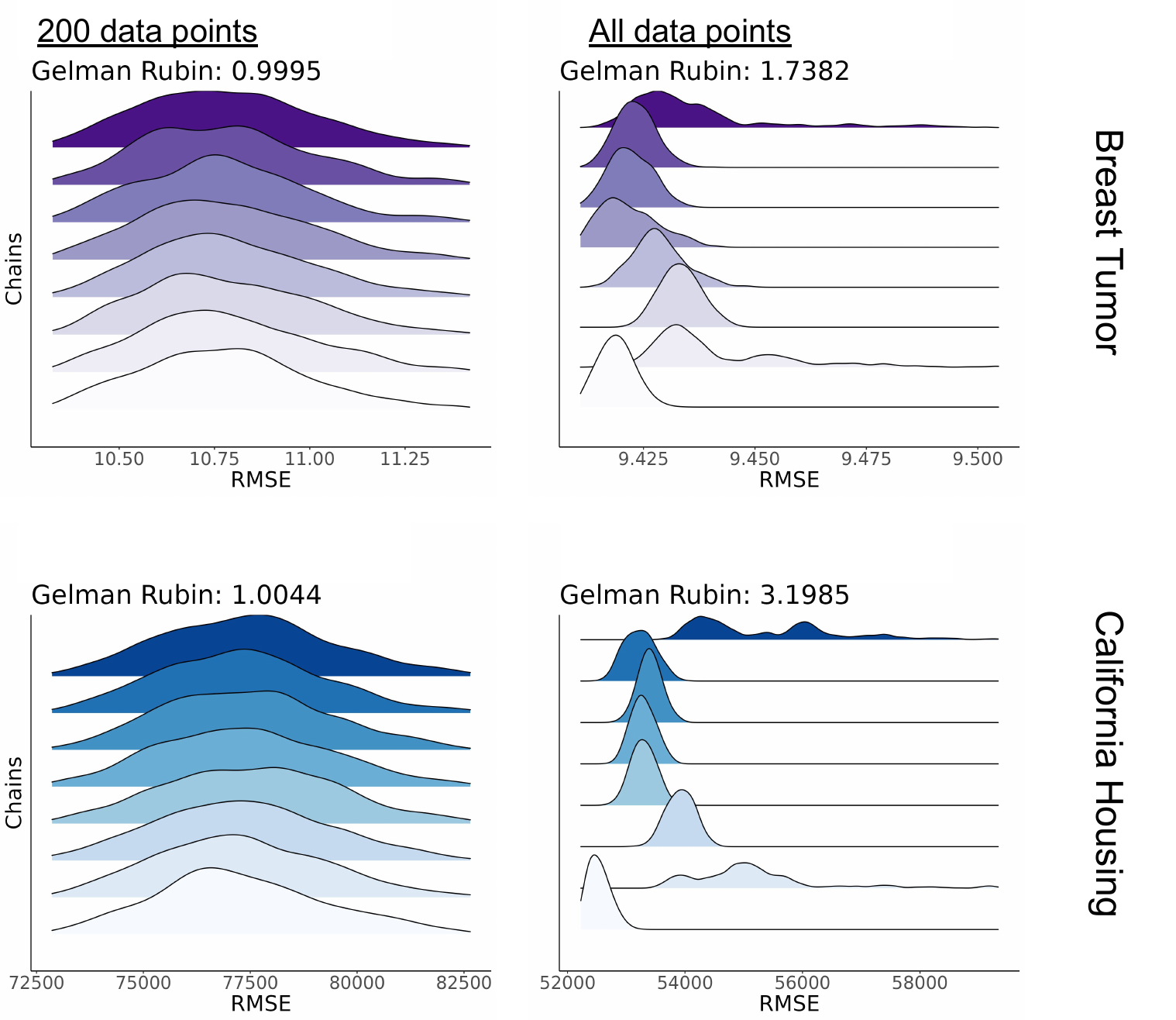}
        \caption{Kernel density plots of BART RMSE values  across different chains and data sets and sample sizes. The number of data points used increases from the left to right column. Different rows correspond to different datasets.}
    \label{fig:bart_rmse}
\end{figure}

\paragraph{Results:}
Across 20 experimental replicates, we compute the GR diagnostic value for BART at each configuration described previously, and summarize our results in \cref{fig:gr_n_p_bart}.
For comparison, we also display the corresponding results for the simplified BART algorithm that we analyzed theoretically.
We observe that the mixing time increases with $n$ for both BART and simplified BART.
Furthermore, even when running the chain with more burn-in samples than recommended, we observe that BART fails to mix on all of the original datasets.
This failure to mix for large $n$ is also visually demonstrated for BART and \sbart respectively in \cref{fig:bart_rmse,,sec:sbart_rmse}, through RMSE density plots.
As $n$ increases the RMSE densities show visible differences. Observing such differences is extremely unlikely if these chains are comprised of samples from a distribution that concentrates around a fixed function.
Furthermore, these plots illustrate a subtle point that mixing reflects the relative magnitude of the between-chain variation compared with the within-chain variation, and not its absolute magnitude. In particular, the between-chain variation for the Breast Tumor dataset is no more than 0.1\% of the RMSE value, while the GR value is 1.74. 
\cref{sec:cumsum} visualizes the same information using the cusum path plots \parencite{yu1998looking}, which have also been used for MCMC diagnostics. We observe that the chains follow a smoother path as the number of data points increases, which is an indication of slow mixing (see \cref{sec:cumsum} for more details).

\subsection{Experiment 2: Root split stuck on suboptimal feature}
\label{sec:expr2}

\paragraph{Additional set-up:}
In contrast to the previous experiment, we study Bayesian CART instead of BART, in order to understand to what extent the two additional moves, \textbf{Change} and \textbf{Swap}, can help avoid bottlenecks.
For each of the 6000 iterations of each chain (including the burn-in iterations), we record the index of the feature on which the root node splits.

\paragraph{Results:}

The results suggest that the root node changes in only a tiny fraction of the iterations, which decreases as $n$ increases. Specifically, when each full dataset is used, the root split changes in less than 0.2\% of the samples on average across 160 chains.
\cref{sec:root_reverse} plots the number of iterations during which the feature on which the root node splits changes (via a prune, change, or swap move) against the number of data points, and demonstrates that the probability of such change decreases with $n$.
Therefore, even with the full set of moves, changing the root split remains a bottleneck for the chain.

\begin{figure}[H]
    \centering
    \includegraphics[scale=0.5]{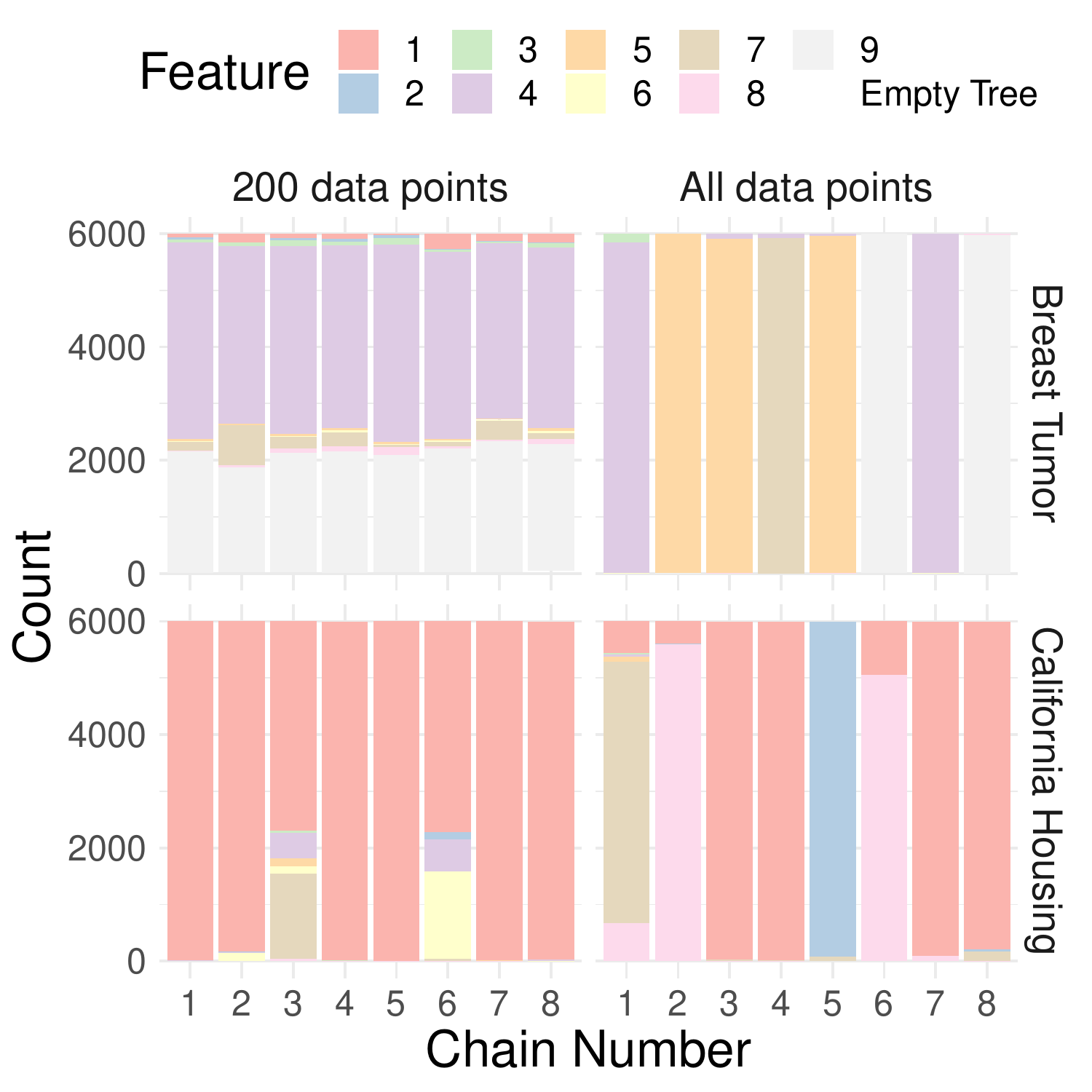}
    \caption{
    Number of iterations on which the root node splits on each feature for different chains on two datasets.
    When $n$ is large (right column), almost all MCMC samples for a single chain of \bcart~have a root split on the same feature, whose index varies across chains.
    The different colors correspond to the different features, and empty tree refers to a single leaf (no splits).
    }
    \label{fig:features_chains}
\end{figure}

In \cref{fig:features_chains}, we summarize the recorded split indices by counting the number of iterations on which the root node splits on each feature.
In each subplot, we display the results separately for each of the 8 independent chains.
We observe that for full datasets, an overwhelming majority of the root splits occur on the same feature, and furthermore, this feature is different for different chains. Similar analysis for \sbart is deffered to \cref{sec:features_chains_simp}.

This further supports our claim about the bottleneck, while demonstrating the instability of the algorithm at the level of tree structures.
Moreover, it implies that the chain can get stuck with a root split that is suboptimal.

\section{Discussion}
\label{sec:discussion}

Poor mixing of the BART and \bcart~MCMC algorithms has been observed since these algorithms were introduced.
In this work we provide the first theoretical analysis that proves a mixing time lower bound scaling with the number of training data points for a simplified version of the BART MCMC algorithm.
Our analysis of \sbart can partially explain why this occurs:
The first split in the tree forms a bottleneck for the mixing of the chain, and the bottleneck becomes more difficult to traverse as the number of training data points increases.
While we do not provide theoretical results for BART, our simulations show that the trend of mixing degradation with the number of data points suggested by our theory also holds for BART. 
We also learn from our simulations that Bayesian CART also encounters a bottleneck when splitting on the first feature; the restricted move set of simplified BART seems not solely responsible for this bottleneck. 

The bottleneck formed by the first split in the tree helps explain one cause of the poor mixing of Bayesian CART and possibly BART.
This split, which may be sub-optimal, forms a bottleneck for the chain and makes it difficult for it to transition to a tree with a differing root split.
Furthermore, both Bayesian CART and BART choose the proposed feature and threshold uniformly at random, and are not guaranteed to make an optimal split initially.

\paragraph{Future Directions:}
We have made four simplifying assumptions to make the theoretical analysis of BART tractable.
Most significantly, we studied a single tree instead of a sum, and only allowed \textbf{Grow} and \textbf{Prune} moves for the MCMC proposal (thereby excluding the \textbf{Change} and \textbf{Swap} moves).
If either \textbf{Change} or \textbf{Swap} moves are allowed, the conductance computations would become more complicated and we may not be able to use the same bottleneck set used in the proof of Theorem \ref{thm:main_thm}.
In future work, we aim to expand our result to hold for a single tree with an unrestricted set of MCMC moves.
Analyzing the behavior of a sum of trees would require developing a new set of analytical tools, and appears prima facie to be more difficult.
Despite these challenges, a result in this vein would be insightful as it would offer a better understanding of the BART algorithm as used in practice.

The remaining simplifying assumptions are mostly technical, and do not greatly affect the generality of our results.
Our assumption of discrete features is relatively benign as a continuous feature can be approximated by its discretization on a sufficiently fine grid. 
Relaxing this assumption would entangle the prior and the likelihood, since in practice for continuous features, splits are selected from a grid of values dependent on the data.
Lastly, our assumption that the variance of the Gaussian prior placed on the leaf parameters, $\sigma^2$, is fixed is not divergent from practice because it converges to a fixed value within a small fraction of the iterations that the algorithm runs for \parencite{chipman2010bart}.
Allowing $\sigma^2$ to be random, as is done in practice, would complicate the form of the integrated likelihood and make an equivalent version of \cref{lma:lratio_impurity} more difficult to prove.

We believe our work should be viewed as only a first step towards rigorously understanding the mixing properties of BART and suggesting principled improvements to this class of MCMC algorithms.
Our work suggests several natural ways to improve the mixing of BART in practice:
The proposal distribution for selecting splits should be data-adaptive rather than uniform at random, the chain should be initialized at an intelligent guess for the possible trees, and the number of MCMC chains run for BART should increase with the number of training data points.
In our future work, we plan to relax the assumptions on moves in simplified BART and develop concrete methods for data-driven splits, tree structure initialization and scaling the number of chains depending on the number of training data points.

\subsection*{Acknowledgements}

We would like to thank the Statistical Computing Facility (SCF) of the Department of Statistics at University of California, Berkeley for computing support, and in particular Professor Jacob Steinhardt for access to his group's large-memory nodes within the SCF Linux cluster.\\
We gratefully acknowledge partial support from NSF TRIPODS Grant 1740855, DMS-1613002, 1953191, 2015341, 2209975, IIS 1741340, ONR grant N00014-17-1-2176, NSF grant 2023505 on Collaborative Research: Foundations of Data Science Institute (FODSI), the NSF and the Simons Foundation for the Collaboration on the Theoretical Foundations of Deep Learning through awards DMS-2031883 and 814639, and a Weill Neurohub grant.
YT was partially supported by NUS Start-up Grant A-8000448-00-00.
\clearpage

\printbibliography

\onecolumn

\appendix

\section{Appendix}
\subsection{Proof details for \cref{sec:lb}}
\subsubsection{Proof of Lemma 3 (Conductance and likelihood ratio)}

\begin{proof}
Let $(v, \tau)$ be the feature and threshold on which the root node splits in tree $\tree$.
Let $S = S(v,\tau)$ denote the set of tree structures whose root node splits on $(v,\tau)$.
Then by assumption, we have $\tree \in S$ and $\tree', \tree_0 \in S^c$, where $\tree_0$ is the empty tree with only a root node.
Assume for now that $\pi(S) \leq \frac{1}{2}$.

Because all possible moves for the Markov chain either add or remove a single terminal split, it is easy to see that $S$ and $S^c$ are connected by a single edge between $\tree_0$ and $\tree''$, the tree of depth 1 whose root node splits on $(v,\tau)$.

The conductance of $S$ can thus be written as
\begin{align}\label{eqn:conductance_of_S}
    \Phi(S) = \frac{\pi(\tree_0)Q(\tree''|\tree_0)}{\pi(S)}.
\end{align}

Let $\psi(-|-)$ denote the transition kernel for the proposal step of the Metropolis-Hastings chain.
We then have $\psi(\tree''|\tree_0) = \frac{1}{2\ncat d}$ (this comes from the probability of choosing a grow move multiplied by the probability of choosing the split $(v,d)$.)
The Metropolis-Hastings filter $\alpha(\tree''|\tree_0)$ is chosen precisely so that
\begin{align*}
    \pi(\tree_0)Q(\tree''|\tree_0) & = \pi(\tree_0)\psi(\tree''|\tree_0)\alpha(\tree''|\tree_0) \\
    & = \min\braces*{\pi(\tree_0)\psi(\tree''|\tree_0), \pi(\tree'')\psi(\tree_0|\tree'')}.
\end{align*}
Plugging in the formula for $\psi(\tree''|\tree_0)$ and continuing the equation gives
\begin{align*}
    \pi(\tree_0)Q(\tree''|\tree_0) \leq \frac{\pi(\tree_0)}{2md}.
\end{align*}

Next, notice that
\begin{align*}
    \pi(S) & = \sum_{\tilde\tree \in S} \pi(\tilde\tree) \geq \pi(\tree).
\end{align*}
As such, the conductance is bounded by a constant factor of the ratio of posterior probabilities as follows:
\begin{equation} \label{eq:conductance_likelihood_intermediate}
    \Phi(S) \leq \frac{\pi(\tree_0)}{2\ncat d\pi(\tree)}.
\end{equation}
Since the posterior probability for any tree $\tilde\tree$ can be written as
$$
p(\tilde\tree | \bX, \by) = \frac{p(\tilde\tree)p(\by|\tilde\tree, \bX)}{\sum_{\check\tree \in \Omega}p(\check\tree)p(\by|\check\tree, \bX)},
$$
the right hand side of \eqref{eq:conductance_likelihood_intermediate} is equal to
\begin{equation*}
    \frac{p(\tree_0)}{2mdp(\tree)}\frac{p(\by|\tree_0, \bX)}{p(\by|\tree, \bX)}.
\end{equation*}
Since prior probabilities depend only on $\alpha$, $\beta$, $m$, and $d$, the desired conclusion follows under the assumption that $\pi(S) \leq \frac{1}{2}$.
If $\pi(S) > \frac{1}{2}$, we repeat the same string of calculations but with $S$ replaced by $S^c$ in \eqref{eqn:conductance_of_S} and with $\tree$ replaced by $\tree'$ thereafter.

\end{proof}

\newpage
\subsubsection{Proof of Lemma 4 (Likelihood ratio and impurity decrease)}

\begin{proof}
We first introduce some notation.
Give the leaves of $\tree$ an ordering, and denote them using $L_1,L_2,\ldots,L_b$.
For $i=1,\ldots,\nleaves$, let $I_i$ denote the set of indices of the data points with $\bx_j \in L_i$.
Let $n_i$ denote the count of data points in leaf $i$, i.e. $n_i := | I_i|$.
Let $\tree(x)$ denote the leaf node that contains an observation $x$, i.e. $\tree(x_i) := L_j$ if $i \in I_j$.
Finally, we will use $C$ to denote absolute constants (independent of both the regression function $f_0$ and the tree $\tree$) that may vary from line to line.

\textit{Step 1: Likelihood ratio to impurity decrease.}
By equation (42) in \parencite{murphy2007conjugate}, we have
\begin{align*}
    \frac{\lklhd(\by|\emptytree, \bX)}{\lklhd(\by|\tree, \bX)} & = \sqrt{\frac{\prod_{i=1}^\nleaves(n_i\sigma^2+a\sigma^2)}{(n\sigma^2+a\sigma^2)(a\sigma^2)^{b-1}}} \exp\braces*{\frac{\sigma^2}{2a\sigma^2(n\sigma^2+a\sigma^2)}\cdot Z^2 - \sum_{i=1}^b \frac{\sigma^2}{2a\sigma^2(n_i\sigma^2+a\sigma^2)}Z_i^2} \\
    & = \sqrt{\frac{\prod_{i=1}^\nleaves(n_i+a)}{(n+a)a^{\nleaves-1}}}\exp\braces*{\frac{1}{2a\sigma^2}\paren*{\frac{1}{n+a}Z^2 - \sum_{i=1}^b \frac{1}{n_i+a}Z_i^2}}
\end{align*}
where $Z = \sum_{j=1}^n y_j$ $Z_i = \sum_{j \in I_j} y_j$ for $i = 1,\ldots,\nleaves$.
We would like to convert the expression in the exponential into an impurity decrease by removing the appearance of $a$ in the denominators.
To do so, we compute the bound:
\begin{align} \label{eq:step1_bound}
    \abs*{\paren*{\frac{1}{n+a}Z^2 - \sum_{i=1}^b \frac{1}{n_i+a}Z_i^2} - \paren*{\frac{1}{n}Z^2 - \sum_{i=1}^b \frac{1}{n_i}Z_i^2}} & = \abs*{\frac{a}{n(n+a)}Z^2 - \sum_{i=1}^b\frac{a}{n_i(n_i+a)}Z_i^2} \nonumber\\
    & \leq \frac{an^2K^2}{n(n+a)} + \sum_{i=1}^b \frac{an_i^2K^2}{n_i(n_i+a)} \nonumber\\
    & \leq a(b+1)K^2.
\end{align}

Note that this is indeed an impurity decrease as
\begin{align*}
    \frac{1}{n}Z^2 - \sum_{i=1}^b \frac{1}{n_i}Z_i^2 & = \frac{1}{n}Z^2 - \sum_{j=1}^n y_j^2 - \sum_{i=1}^b \paren*{\frac{1}{n_i}Z_i^2 - \sum_{\bx_j \in L_i} y_j^2}\\
    & = -\sum_{j=1}^n(y_j - \bar{y})^2 + \sum_{i=1}^b \sum_{\bx_j \in L_i} (y_j - \bar{y}_{L_i})^2,
\end{align*}
where $\bar{y}_{L_i} = n_i^{-1}Z_i$.

\textit{Step 2: Concentration of impurity decrease conditioned on counts.}
Condition on $n_1,\ldots,n_b$ and denote $\tilde q_i = \frac{n_i}{n}$ for $i=1,\ldots,b$.
The impurity decrease may be written as a variance:
\begin{align*}
    -\frac{1}{n}Z^2 + \sum_{i=1}^b \frac{1}{n_i}Z_i^2 & = -n\paren*{\sum_{i=1}^b\tilde q_i \bar {y}_{L_i}^2 - \paren*{\sum_{i=1}^b \tilde q_i\bar y_{L_i}}^2}. 
\end{align*}
We may rewrite the expression in parentheses on the right in matrix form:
\begin{align*}
    \sum_{i=1}^b\tilde q_i \bar {y}_{L_i}^2 - \paren*{\sum_{i=1}^b \tilde q_i\bar y_{L_i}}^2 = \bw^T\paren*{\bI_b -\sqrt{\tilde\bq}\sqrt{\tilde\bq}^T} \bw
\end{align*}
where $w_i = \sqrt{\tilde q_i}\bar y_{L_i}$ for $i = 1,\ldots, b$.

Denote
\begin{equation*}
    \tilde\Delta = \sum_{i=1}^b\tilde q_i \E\braces{f_0(\bx)|\bx \in L_i}^2 - \paren*{\sum_{i=1}^b \tilde q_i\E\braces{f_0(\bx)|\bx \in L_i}}^2,
\end{equation*}
and notice that we may also write
\begin{equation*}
    \tilde\Delta = \E\braces{\bw}^T\paren*{\bI_b -\sqrt{\tilde\bq}\sqrt{\tilde\bq}^T}\E\braces{\bw}.
\end{equation*}
We therefore have
\begin{align} \label{eq:impurity_dec_conc_step2}
    & \paren*{\sum_{i=1}^b\tilde q_i \bar {y}_{L_i}^2 - \paren*{\sum_{i=1}^b \tilde q_i\bar y_{L_i}}^2} - \tilde \Delta \nonumber \\
    = ~& (\bw -\E\braces{\bw})^T\paren*{\bI_b -\sqrt{\tilde\bq}\sqrt{\tilde\bq}^T} (\bw -\E\braces{\bw}) + 2\paren*{\bw -\E\braces{\bw}}^T \paren*{\bI_b -\sqrt{\tilde\bq}\sqrt{\tilde\bq}^T} \E\braces{\bw}.
\end{align}
We shall bound these two terms separately.
To this end, we compute
\begin{align*}
    \norm*{\bI_b -\sqrt{\tilde\bq}\sqrt{\tilde\bq}^T}_F^2 & = \norm*{\bI_b}_F^2 - 2\text{Tr}\paren*{\sqrt{\tilde\bq}^T \bI_b\sqrt{\tilde\bq}} + \norm*{\sqrt{\tilde\bq}\sqrt{\tilde\bq^T}}_F^2 \\
    & = b - 1.
\end{align*}
Similarly,
\begin{align*}
    \norm{\bI_b -\sqrt{\tilde\bq}\sqrt{\tilde\bq}^T}_{op} = 1.
\end{align*}
Note also that, by Hoeffding's lemma \parencite{vershynin2018high}, $\bw$ has independent sub-Gaussian entries, each with squared sub-Gaussian norm at most
\begin{align*}
    \norm*{\sqrt{\tilde q_i}\bar y_{L_i}}_{\psi_2}^2 \leq \frac{\tilde q_i K^2}{n_i} = \frac{K^2}{n}.
\end{align*}
Applying the Hanson-Wright inequality \parencite{vershynin2018high}, we therefore get
\begin{align} \label{eq:hanson_wright}
    (\bw -\E\braces{\bw})^T\paren*{\bI_b -\sqrt{\tilde\bq}\sqrt{\tilde\bq}^T} (\bw -\E\braces{\bw}) \leq \frac{CbK^2\sqrt{\log n}}{n}
\end{align}
with probability at least $1 - 1/4n$.

Meanwhile, we have
\begin{align*}
    \norm*{\E\braces{\bw}}_2^2 & = \sum_{i=1}^b \tilde q_i \paren*{\E\braces*{f_0(\bx)|\bx \in L_i}}^2 \\
    & \leq K^2.
\end{align*}
Hence, using Hoeffding's inequality, we have
\begin{align} \label{eq:cross_term_bound}
    \paren*{\bw -\E\braces{\bw}}^T \paren*{\bI_b -\sqrt{\tilde\bq}\sqrt{\tilde\bq}^T} \E\braces{\bw} \leq CK^2\sqrt{\frac{\log n}{n}}
\end{align}
with probability at least $1-4/n$.
Plugging \eqref{eq:hanson_wright} and \eqref{eq:cross_term_bound} into \eqref{eq:impurity_dec_conc_step2} gives us
\begin{equation}
    \paren*{\sum_{i=1}^b\tilde q_i \bar {y}_{L_i}^2 - \paren*{\sum_{i=1}^b \tilde q_i\bar y_{L_i}}^2} - \tilde \Delta \leq CK^2\sqrt{\log n} \paren*{\frac{b}{n} + \frac{1}{\sqrt{n}}}
\end{equation}
with probability at least $1 - 2/n$.

\textit{Step 3: Decondition on counts.}
Using the law of total variance, write
\begin{align*}
    \Delta & = \Var\braces{f_0(\bx)} - \E\braces{\Var\braces{f_0(\bx)|\tree(\bx)}} \\
    & = \Var\braces*{\E\braces{f_0(\bx)~|~\tree(\bx)}} \\
    & = \sum_{i=1}^b q_i \E\braces{f_0(\bx)~|~\bx \in L_i}^2 - \paren*{\sum_{i=1}^b q_i \E\braces{f_0(\bx)~|~\bx \in L_i}}^2
\end{align*}
where $q_i = \P\braces*{y \in L_i}$ for $i = 1,\ldots,b$.
Since $n\tilde \bq$ is a multinomial distribution with parameters $n$ and $\bq$, we may apply Hoeffding's inequality two more times to get
\begin{equation*}
    \abs*{\sum_{i=1} \paren*{\tilde q_i - q_i} \E\braces{f_0(\bx)~|~\bx \in L_i}} \leq CK\sqrt{\frac{\log n}{n}}
\end{equation*}
and
\begin{equation} \label{eq:bound_for_1st_term_diff}
    \abs*{\sum_{i=1} \paren*{\tilde q_i - q_i} \E\braces{f_0(\bx)~|~\bx \in L_i}^2} \leq CK^2\sqrt{\frac{\log n}{n}}
\end{equation}
jointly with probability at least $1-1/2n$.
On this event, we also have
\begin{align} \label{eq:bound_for_2nd_term_diff}
    & \abs*{\paren*{\sum_{i=1}^b \tilde{q}_i \E\braces{f_0(\bx)~|~\bx \in L_i}}^2 - \paren*{\sum_{i=1}^b q_i \E\braces{f_0(\bx)~|~\bx \in L_i}}^2} \nonumber\\
    = ~& \abs*{\sum_{i=1}^b \paren*{\tilde{q}_i  - q_i}\E\braces{f_0(\bx)~|~\bx \in L_i}} \cdot \abs*{\sum_{i=1}^b \paren*{\tilde{q}_i + q_i} \E\braces{f_0(\bx)~|~\bx \in L_i}} \nonumber\\
    & \leq 2K\abs*{\sum_{i=1}^b \paren*{\tilde{q}_i  - q_i}\E\braces{f_0(\bx)~|~\bx \in L_i}} \nonumber\\
    & \leq CK^2\sqrt{\frac{\log n}{n}}.
\end{align}
Using \eqref{eq:bound_for_1st_term_diff} and \eqref{eq:bound_for_2nd_term_diff}, we get
\begin{align*}
    \abs*{\tilde \Delta - \Delta} \leq CK^2\sqrt{\frac{\log n}{n}}.
\end{align*}

Conditioning on this event and further conditioning on the $1-1/2n$ event guaranteeing \eqref{eq:impurity_dec_conc_step2} then gives us:
\begin{align*}
    \abs*{\paren*{\frac{1}{n+a}Z^2 - \sum_{i=1}^b \frac{1}{n_i+a}Z_i^2} + n\Delta} & \leq a(b+1)K^2 + CK^2\sqrt{\log n} \paren*{b + \sqrt{n}} + CK^2\sqrt{n\log n} \\
    & \leq CK^2\sqrt{\log n}\paren*{\sqrt{n} + b}
\end{align*}
since it only makes sense to choose $a \leq 1$.

Finally, we have
\begin{align*}
    \log \sqrt{\frac{\prod_{i=1}^\nleaves(n_i+a)}{(n+a)a^{\nleaves-1}}} & = \frac{1}{2}\sum_{i=1}^b \log(n_i+a) - \frac{1}{2}\log(n+a) - (b-1)\log a \\
    & \leq b\log(n/a).
\end{align*}
\end{proof}

\subsection{Additional Simulation Results}
\subsubsection{Simplified BART RMSE Kernel Density}\label{sec:sbart_rmse}

\begin{figure}[H]
    \centering
    \includegraphics[scale=0.5]{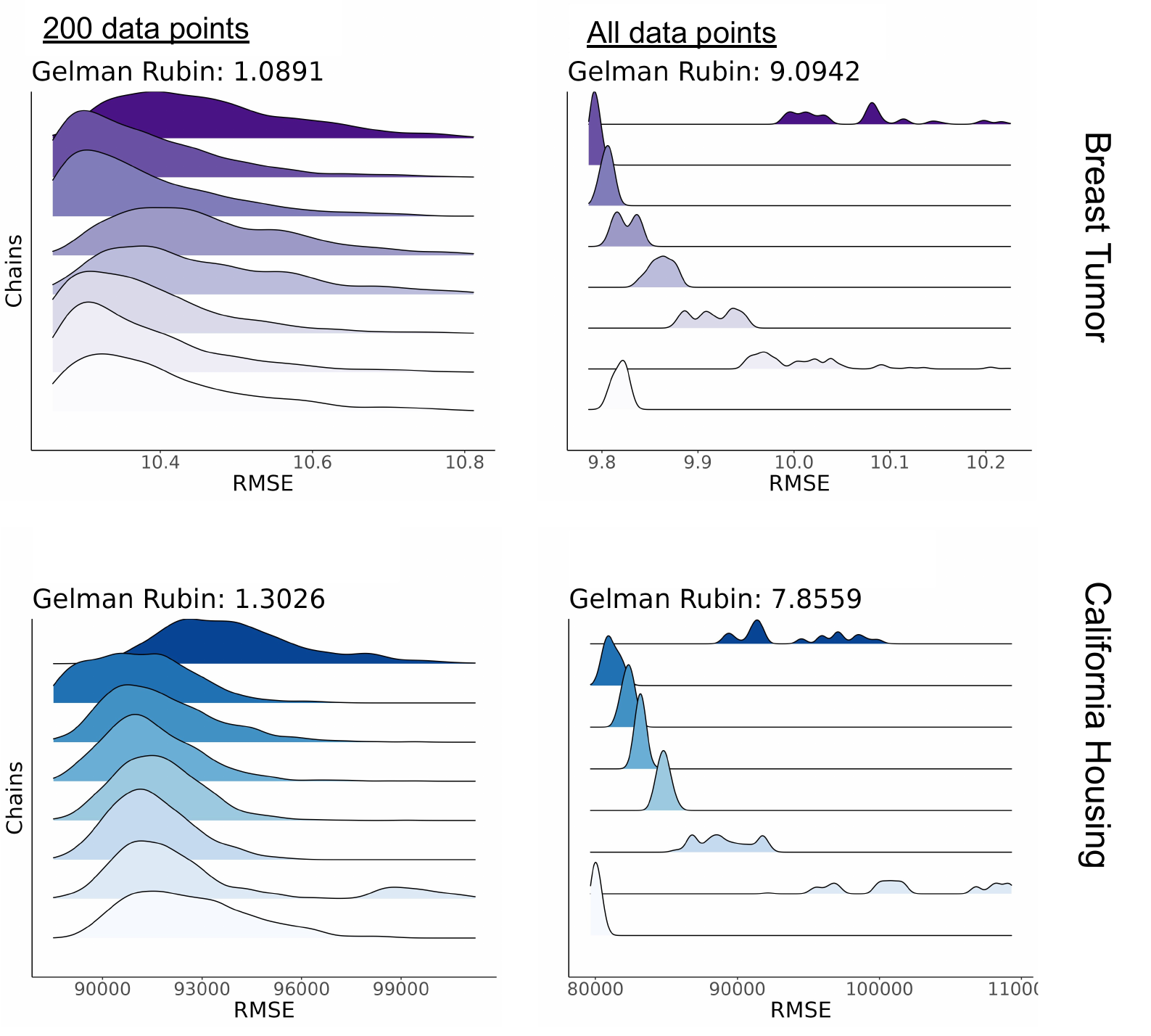}
        \caption{Kernel density plots of \sbart RMSE values  across different chains and data sets and sample sizes. The number of data points used increases from the left to right column. Different rows correspond to different datasets.}
    \label{fig:b-cart_rmse}
\end{figure}

\subsubsection{Cusum Diagnostics for BART and \sbart}\label{sec:cumsum}
In this section we visualize the BART and \sbart mixing using the cusum path plot proposed by \cite{yu1998looking}. We now describe the cusum plot formally: Suppose we have $L$ samples from $J$ different MCMC chains obtained after a burn-in period. Let $x_{ij}$ denote the RMSE value of the $i$th sample from the $j$th chain, $\bar x_j$ denote the mean RMSE sample from the $j$th chain, and let $S^{(j)}_t = \sum_{i=1}^t (x_{ij} -  \bar x_j)$ be the cumulative sum of deviations from the mean. In a cusum plot, we plot $S^{(j)}_t$ against $t$. 
A "hairy" cusum plot, in which the cumulative sum varies randomly around a mean of zero, represents fast mixing. A smooth cusum plot, represents shifts in the sampling process, and is an indicating of slow mixing. In our case, a smooth cusum path would translate into a sub-sequence of tree functions within a given chain, whose RMSE value is higher or lower than the chain average.

\begin{figure}[H]
    \centering
    \includegraphics[scale=0.4]{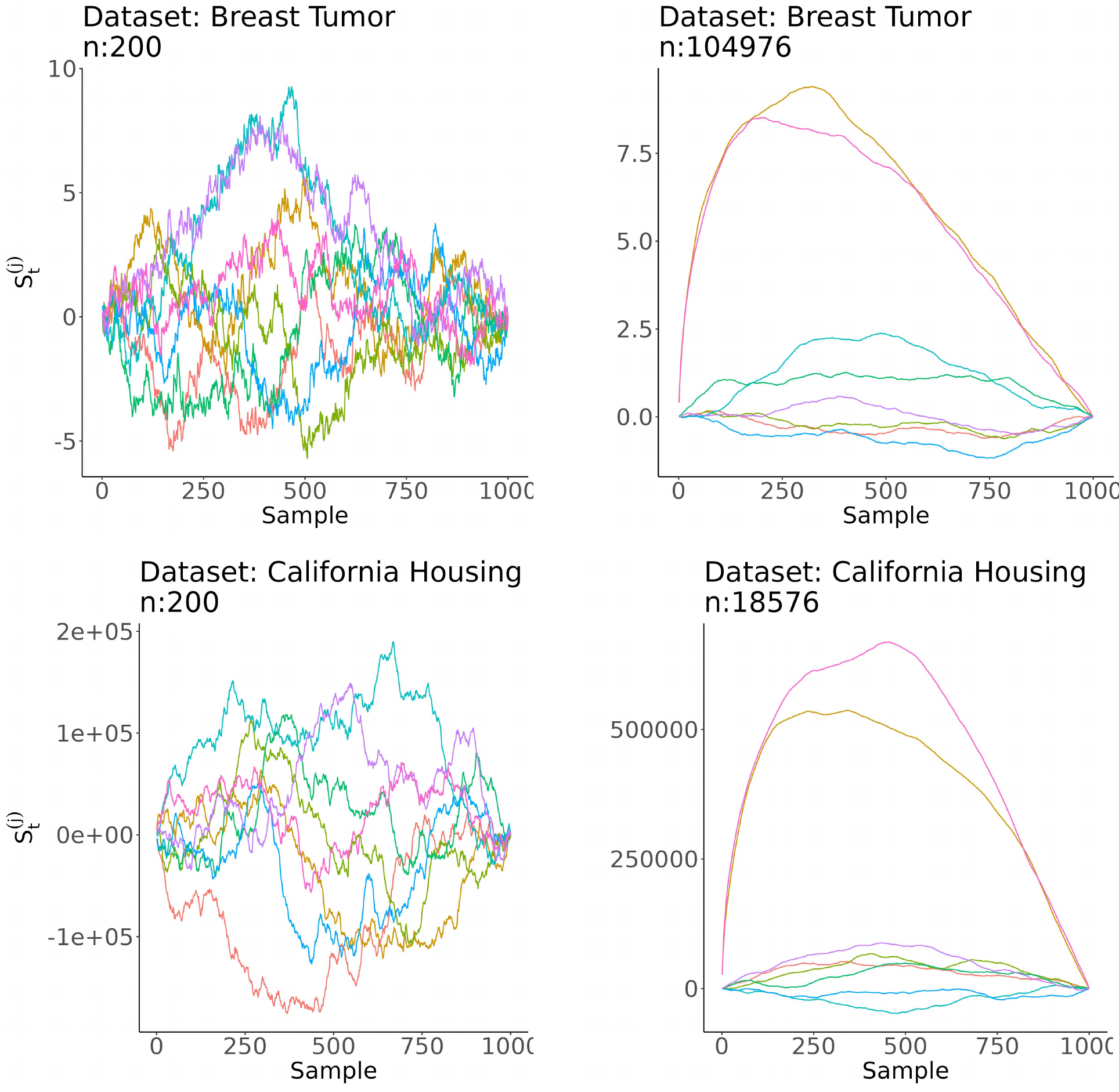}
        \caption{BART cusum path plot become smoother as more data points are used. $S^{(j)}_t$ is plotted against $t$ for 8 different chains. The number of data points used increases from the left to right column. Different rows correspond to different datasets, and different colors correspond to different chains.}
    \label{fig:cumsum-bart}
\end{figure}

\begin{figure}[H]
    \centering
    \includegraphics[scale=0.4]{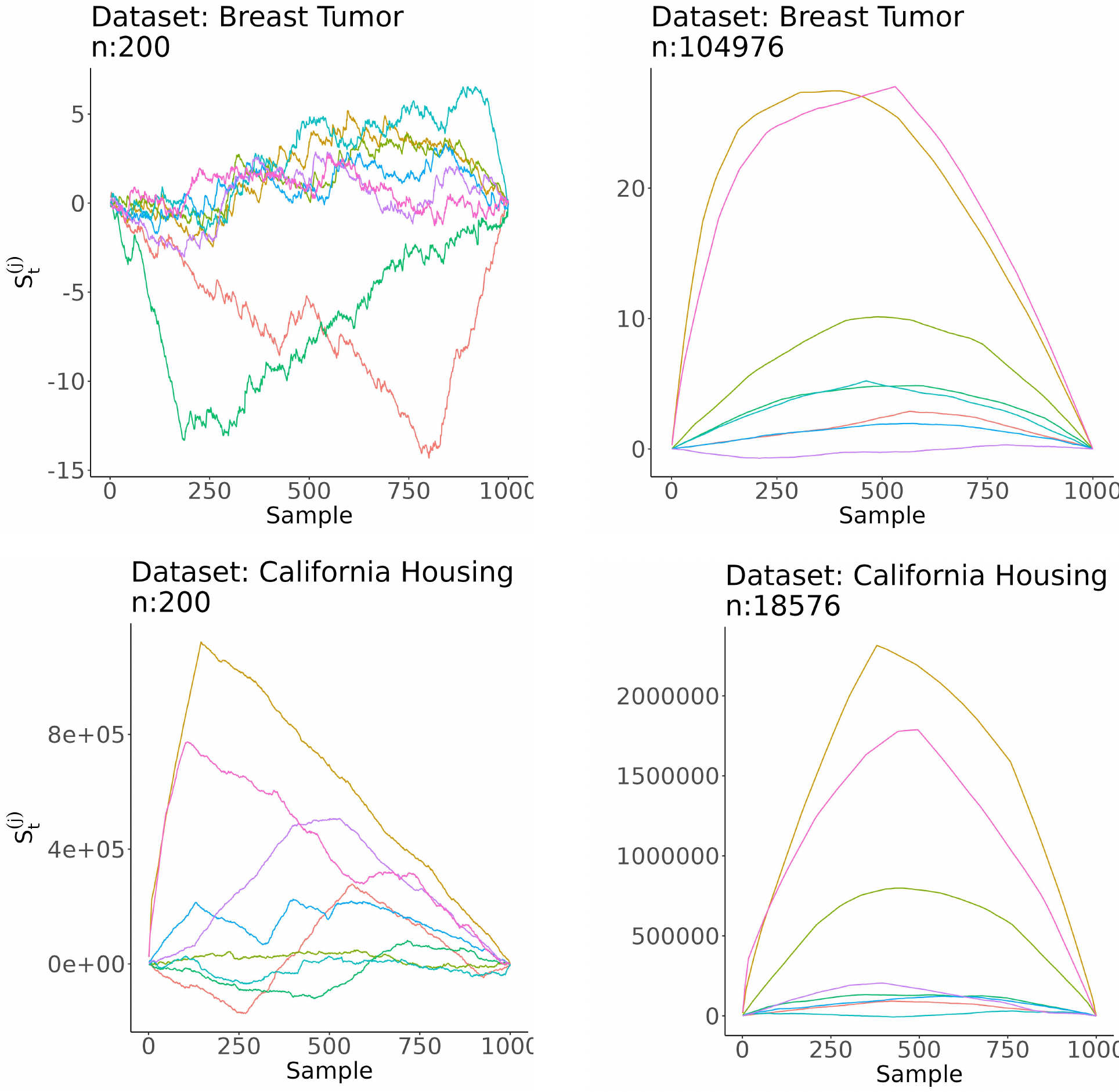}
        \caption{Simplified BART cusum path plot become smoother as more data points are used. $S^{(j)}_t$ is plotted against $t$ for 8 different chains. The number of data points used increases from the left to right column. Different rows correspond to different datasets, and different colors correspond to different chains.}
    \label{fig:cumsum-sbart}
\end{figure}

\cref{fig:cumsum-bart,,fig:cumsum-sbart} show cusum plot for BART and \sbart respectively, across 8 different chains.

\subsection{Reversing the Root Split}\label{sec:root_reverse}
\begin{figure}[H]
    \centering
    \includegraphics[scale=0.6]{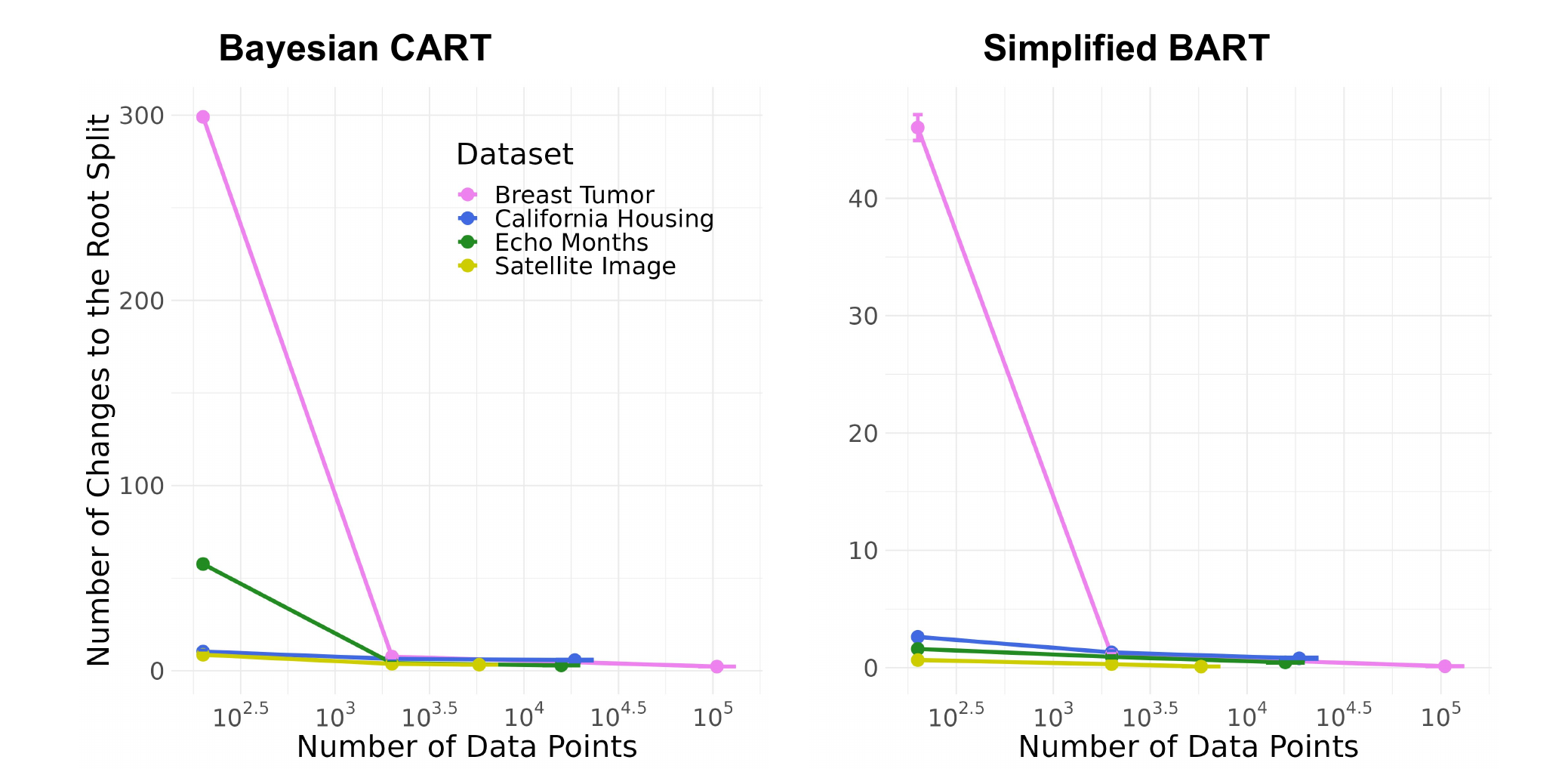}
    \caption{The number of MCMC iterations during which the feature on which the root node splits changes is small, and decreases as the number of training data points increases. Right panel shows the results for B-CART and left for \sbart.
    Error bars are calculated over 160 different runs using the same training data.}
    \label{fig:first_split}
\end{figure}

\subsubsection{Simplified BART Root Node Split Indices}\label{sec:features_chains_simp}

\begin{figure}[H]
    \centering
    \includegraphics[scale=0.5]{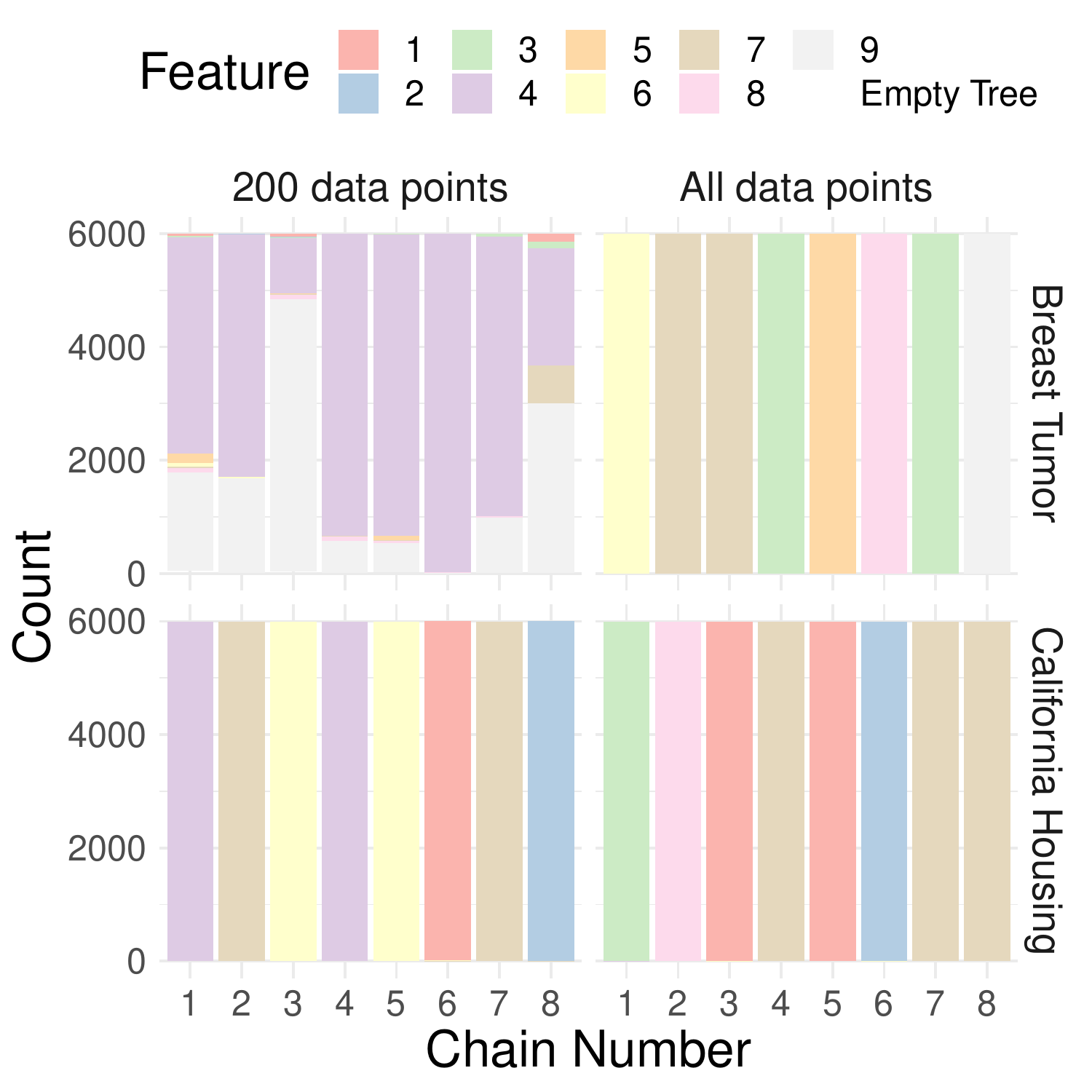}
    \caption{Number of iterations on which the root node splits on each feature for different chains on several datasets.
    When $n$ is large, almost all MCMC samples for a single chain of \sbart have a root split on the same feature, whose index varies across chains. The different colors correspond to the different features, and empty tree refers to a single leaf (no splits).}
    \label{fig:features_chains_simp}
\end{figure}

\end{document}